\documentclass{fairmeta} 

\usepackage{latexsym}
\usepackage{enumitem}      
\usepackage{amsmath}       
\usepackage{amssymb}       
\usepackage{algorithm}     
\usepackage{algorithmic}   
\usepackage{array}         
\usepackage{wrapfig}       
\usepackage{colortbl}      
\usepackage{xspace}

\definecolor{accenttint}{RGB}{237, 243, 252}  
\definecolor{rankfirst}{HTML}{FFF2CC}
\definecolor{ranksecond}{HTML}{E9EBEE}
\definecolor{rankthird}{HTML}{FBE2CB}
\definecolor{seedtint}{HTML}{EDF6F4}

\definecolor{takeawayframe}{gray}{0.20}
\definecolor{takeawayback}{gray}{0.95}
\newtcolorbox{takeaway}[1][]{%
  enhanced, unbreakable, before skip=12pt, after skip=12pt,
  colback=takeawayback, colframe=takeawayframe,
  colbacktitle=takeawayframe, coltitle=white, fonttitle=\bfseries\normalsize,
  title={Takeaway\ifstrempty{#1}{}{~#1}},
  boxrule=1pt, arc=1.5mm,
  left=2.5mm, right=2.5mm, top=1.5mm, bottom=1.5mm,
  toptitle=1mm, bottomtitle=1mm, fontupper=\normalsize
}

\definecolor{calloutframe}{RGB}{52, 103, 170}
\definecolor{calloutback}{RGB}{235, 242, 255}
\newtcolorbox{callout}[1]{%
  enhanced, before skip=12pt, after skip=12pt,
  colback=calloutback, colframe=calloutframe,
  colbacktitle=calloutframe, coltitle=white, fonttitle=\bfseries,
  halign=left, title={#1}, boxrule=1pt, arc=2.5mm,
  left=2.5mm, right=2.5mm, top=2mm, bottom=2mm,
  toptitle=1mm, bottomtitle=1mm
}

\renewcommand\authorformat[2][]{{\sffamily\bfseries #2\ifstrempty{#1}{}{$^{#1}$}}}
\renewcommand\affiliationformat[2][]{{\normalsize \ifstrempty{#1}{}{$^{#1}$}#2}}

\title{Self-Evolving Skills via Su\underline{r}rogat\underline{e}-Guided \underline{Solve}-and-Reproduce}

\author[1,2]{Jiale Liu}
\author[1,3]{Pinze Ren}
\author[3]{Yuqi Xia}
\author[1]{Huan Wang}
\author[1]{Zhenlin Zhao}
\author[1]{Siming Dong}
\affiliation[1]{Cleer Science}
\affiliation[2]{The University of Edinburgh}
\affiliation[3]{Tsinghua University}
\contribution{\texttt{\{wanghuan, zhenlinzhao, william.d\}@cleerlab.com}}
\date{\today}

\abstract{%
Agent skills are portable packages of instructions and resources an agent consults at deployment. Self-evolving them fails in two ways today. First, skills evolved from scratch underperform human-curated ones and, on a weak model, using no skill at all. Second, an evolution-time pass records one lucky trajectory that a fresh stochastic agent often fails to reproduce at deployment. We present reSolve, a per-task, oracle-in-the-loop framework built on three components. It decouples interactive solving from a self-contained deliverable that is independently re-executed in a fresh container, a protocol we call solve-and-reproduce. It enhances the sparse reward signal with a surrogate verifier that cannot access hidden tests or reference answers. It then runs verifier-guided beam search over a solution-construction graph. Within a fixed harness, a cheap model self-evolves skills that reach $74.9\%$ mean-of-3, $+14.8$ points over the $60.1\%$ human-curated baseline, exceeding the strongest official curated-skill result ($67.3\%$, GPT-5.5/OpenHands). We also report observed failure cases and domain-level results, including performance on the 14 Natural Science tasks, to clarify when the approach does and does not help.
\vspace{0.5cm}
}

\hypersetup{
  pdftitle={Self-Evolving Skills via Surrogate-Guided Solve-and-Reproduce},
  pdfauthor={Cleer Science}
}

\renewcommand{\titlelogos}{%
  \raisebox{-0.5\height}{\includegraphics[height=0.85cm]{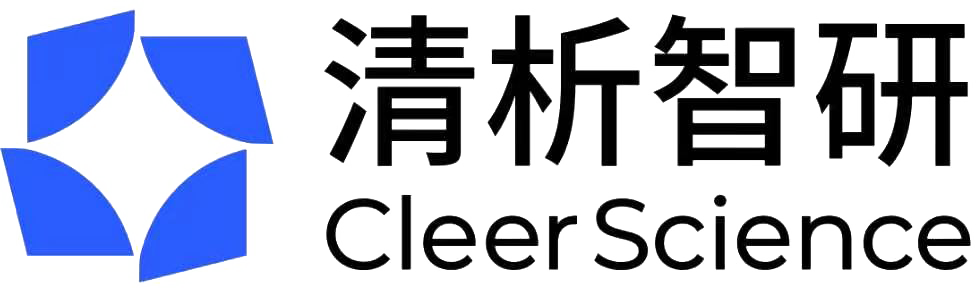}}\hspace{0.5cm}%
  \raisebox{-0.5\height}{\includegraphics[height=0.85cm]{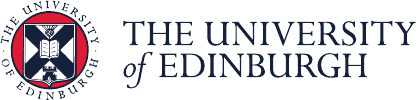}}\hspace{0.5cm}%
  \raisebox{-0.5\height}{\includegraphics[height=0.85cm]{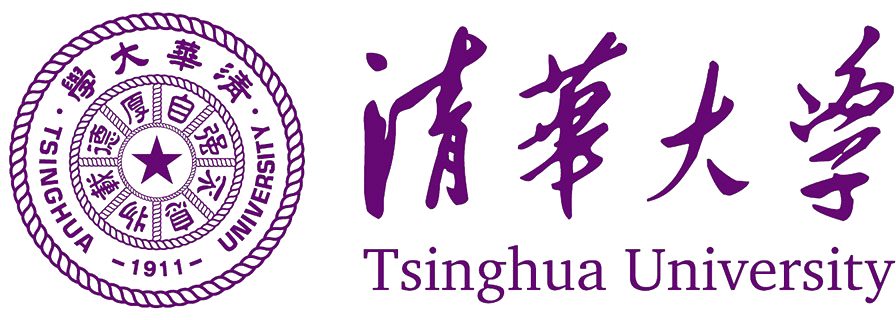}}%
}

\begin{document}
\maketitle

\section{Introduction}
\label{sec:intro}

Agent skills are portable packages of instructions, procedures, and sometimes
executable code that a large language model (LLM) agent reads before acting on a task
\citep{IntroducingAgentSkills}. A skill encodes procedural knowledge that the task requires
but the base model may lack, for example, which files to inspect, what conventions the output must satisfy, or the sequence of operations that avoids known errors. Because a skill is stored as
text and code rather than in model weights, it can be attached to any agent without
retraining, inspected and edited directly, and reused across tasks and models.

Good skills can substantially improve agent performance. In open-ended embodied settings, an
agent that accumulates a library of executable skills obtains $3.3\times$ more unique
items in Minecraft and unlocks milestones up to $15.3\times$ faster than agents without
one \citep{voyager}. On web-navigation benchmarks, supplying agents with workflows
induced from experience improves success rates by $24.6\%$ and $51.1\%$ relative on
Mind2Web and WebArena \citep{awm}, and self-synthesized skill APIs yield gains of similar
magnitude on live websites \citep{skillweaver}. Curated skills likewise raise the task
scores of state-of-the-art coding agents on expert-designed scientific analysis tasks
\citep{scivisskills}. Skill quality is therefore a major determinant of agent
capability, independent of the choice of model.

Producing a good skill, however, requires substantial human effort. The author must solve
the task, distill the solution into steps that a stochastic executor can follow reliably,
and anticipate the ways an agent will misread them. This curation is slow and expensive, and it
must be redone for every task. A natural alternative is to let the agent self-evolve its skills.
When a task provides an automatic checker, the agent can propose candidate skills, test
them against the checker, and keep those that pass
\citep{voyager,skillweaver,coevoskills}. However, two failure modes limit this approach in practice. The first is that, as commonly observed, self-evolved skills can degrade agent performance \citep{sokskills}.
Pure self-evolution falls below not using skills at all on a weak model, because a weak generator-verifier pair could fill the skill library with
plausible but incorrect candidates whose evolution-time passes are search luck.

The second failure mode is subtler and, we argue, more consequential. Current evolution frameworks reward the wrong objective. A self-evolved skill is not itself the solution; it is advice that a stochastic agent re-executes at deployment. An evolution-time pass therefore records a single successful trajectory, and that success need not recur when a fresh agent later reads the skill and re-derives the answer. We call this shortfall the \emph{reproduction gap}. This means, for example, a skill that passes evolution-time checks on a weak model may fail to reproduce on a stronger one, and vice versa. The gap is built into current self-evolution frameworks, which reward a candidate skill for producing a solution trace once, not for reproducing the solution when re-executed.

Therefore, we present reSolve (Self-Evolving Skills via Su\underline{r}rogat\underline{e}-Guided \underline{Solve}-and-Reproduce), a self-evolving framework that mitigates both failure modes (\Cref{fig:arch},
\Cref{sec:method}). For the first, reSolve uses a beam search over a solution-construction
graph to expose search breadth and depth as a test-time-compute parameter, which reduces
the role of luck in evolution. The search is guided by a surrogate verifier that
supplies the dense reward signal, while the grader's full-pass signal gates continuation.
Hidden tests, reference answers, and grader-side diagnostics remain withheld from the
producer. For the second, reSolve changes what evolution certifies
through solve-and-reproduce. It decouples interactive solving from a
self-contained deliverable that is independently re-executed in a fresh container, so that a pass
certifies the frozen deliverable under a new execution rather than the original successful trace. We evaluate reSolve on SkillsBench, a benchmark of 87 tasks that provides an oracle and curated skills for each task. We find that, first, self-evolution is most effective as self-refinement. With a cheap model, reSolve reaches $74.9\%$ mean-of-3, $+14.8$ points over the $60.1\%$ curated baseline, and exceeds the strongest official curated-skill result ($67.3\%$, GPT-5.5/OpenHands), although the leaderboard numbers come from different harnesses. Second, even evolving from scratch, with no curated skill to read, lifts this $26.9\%$-no-skill model to $44.0\%$, within the range that frontier agents attain with no
skill at all, on par with Claude Opus and behind only GPT-5.5.

Our contributions are as follows.

\begin{itemize}[leftmargin=*]
  \item \textbf{C1 (reSolve).} A framework that lets an LLM self-evolve its skills and mitigates both performance degradation and the reproduction gap.
  \item \textbf{C2 (verifier-guided beam search).} reSolve casts per-task evolution as beam
    search over a solution-construction graph. A surrogate verifier
    supplies the dense reward as the heuristic value, and the grader's full-pass signal is the
    stopping gate. This exposes search breadth and depth as a test-time-compute
    parameter and reduces the role of luck in evolution.
  \item \textbf{C3 (solve-and-reproduce).} A protocol that decouples interactive solving from a self-contained
    deliverable, independently re-executed in a fresh container, so that evolution tests
    the frozen deliverable rather than the authoring trace.
  \item \textbf{C4 (empirical study).} On SkillsBench, we characterize when
    self-evolution pays off. It is most effective as self-refinement, while from-scratch
    evolution reaches the no-skill frontier range but stays below the curated baseline. We also report failure cases and discuss the limitations of self-evolution.
\end{itemize}

\section{Related Work}
\label{sec:related}

\paragraph{Agent skills and procedural memory.}
Agent skills externalize procedural knowledge as portable
packages that contain instructions, scripts, and supporting files, which an LLM agent
loads without weight changes \citep{IntroducingAgentSkills,sokskills}.
\citet{voyager} store self-verified executable programs in an
embedding-indexed skill library, and an agent drawing on this library obtains
$3.3\times$ more unique items in Minecraft. \citet{awm} induce reusable
workflows from successful web-navigation trajectories, raising success rates by
up to $51.1\%$ on WebArena. \citet{skillweaver} extend this idea to live
websites by first discovering and practicing procedures, then exposing them as
callable APIs. SkillsBench provides the broadest paired evaluation of this
abstraction. Across 87 tasks and eight domains, curated skills raise the average
pass rate from $33.9\%$ to $50.5\%$, whereas one-shot self-generated skills
fall below the no-skill baseline \citep{skillsbench}. Together, these
studies show that external procedural knowledge can substantially boost agent
capability, but the gains hinge on skill quality. This motivates work on
methods that can reliably construct and improve such artifacts.

\paragraph{Learning and evolving skill artifacts.}
A growing body of work learns reusable skills from agent execution experience at
various timescales. \citet{expel} extract task-level insights from successful
and failed trajectories as transferable rules. \citet{trace2skill} consolidate
trajectories into a static skill directory and evaluate transfer on disjoint
tasks. \citet{skillrl} distill experience into a hierarchical skill bank
co-evolved with an agent policy through reinforcement learning. \citet{skillopt}
cast a single skill document as trainable external state, applying bounded text
edits from rollout batches and accepting an edit only when it improves a
held-out validation score. \citet{skillaudit} eliminate the need for
ground-truth labels by contrasting paired with-skill and without-skill
trajectories. \citet{coevoskills} couple skill evolution with surrogate test
co-evolution, querying the hidden oracle only after the surrogate passes. These
methods extract persistent guidance from batches of execution evidence, and they
typically evolve a shared skill library across multiple tasks. reSolve studies a
complementary per-task setting.  It uses beam search and a surrogate verifier to
construct a self-contained deliverable for a single workspace, and certifies the
deliverable through independent re-execution.

\paragraph{Grounded feedback and surrogate verification.}
Iterative self-refinement is a natural strategy for improving LLM outputs, but
its effectiveness depends on the quality of the feedback signal.
\citet{selfrefine} show that language feedback alone can iteratively improve
generation quality, and \citet{reflexion} extend this with verbal reinforcement
from environment traces. \citet{critic} demonstrate that grounding critique in
external tool interactions substantially improves factuality, while
\citet{cannotselfcorrect} show that without such grounding, intrinsic
self-correction can reduce accuracy. Generated executable tests are an
alternative form of grounded signal. \citet{codet} cluster programs by agreement
on LLM-generated tests and use dual execution agreement for selection.
\citet{coevoskills} apply this principle to agent skills by having an
information-isolated surrogate synthesize tests over produced outputs, which creates a
verification loop independent of the skill generator. The shared conclusion
is that language models refine outputs effectively only when feedback is
anchored in execution or external evidence. reSolve accordingly guides skill
repair with executable surrogate check suites.

\paragraph{Search over language artifacts.}
Rather than committing to a single refinement trajectory, search-based methods
retain and explore multiple candidates under an evaluative signal. \citet{lats}
combine language-model value estimates with Monte Carlo tree search and
environmental feedback for agent decision-making. \citet{gepa} evolve prompts
through trajectory reflection and Pareto-frontier selection. AFlow and
AlphaEvolve search over agent workflows and programs under executable evaluation
\citep{aflow,alphaevolve}. \citet{adas} frame agent architecture design itself
as a search problem, with novel agent configurations generated and evaluated
automatically, and Promptbreeder applies evolutionary self-referential mutation
to prompts \citep{promptbreeder}. Across these systems, explicit search over
language-level artifacts can outperform sequential refinement.

\paragraph{Execution semantics and reproducible deployment.}
Making agent behavior executable improves inspectability but does not by itself
ensure that a skill will produce the same outcome when re-used. CodeAct lets
agents emit and revise executable Python actions \citep{codeact}, and Voyager
self-verifies programs before storing them \citep{voyager}. SkillWeaver
packages practiced web procedures as callable APIs, yet reports deployment
failures from selecting the wrong API or supplying incorrect parameters
\citep{skillweaver}. Existing skill optimizers likewise evaluate through fresh
stochastic rollouts \citep{coevoskills,trace2skill,skillopt}, which estimate an
execution distribution rather than certify an invariant outcome, so one
passing rollout does not guarantee reproduction by a new agent. This distinction between
authoring-time success and deployment-time reliability motivates reSolve's
solve-and-reproduce protocol, which tests a frozen deliverable across a fresh
execution boundary rather than certifying the original authoring trace.

\section{Method}\label{sec:method}

reSolve evolves one task-specific skill package at a time.  The goal is not a
single successful trajectory but a frozen package that a fresh agent can re-use
to solve the same task.  Each search round follows the same sequence.  A producer
agent authors a candidate skill, the candidate is frozen and handed to a new
agent in a fresh container, and the resulting files are evaluated.  A full pass
ends the search.  Otherwise, a surrogate verifier diagnoses defects and guides
the next refinement.

\begin{figure*}[t]
  \centering
  \includegraphics[width=\textwidth]{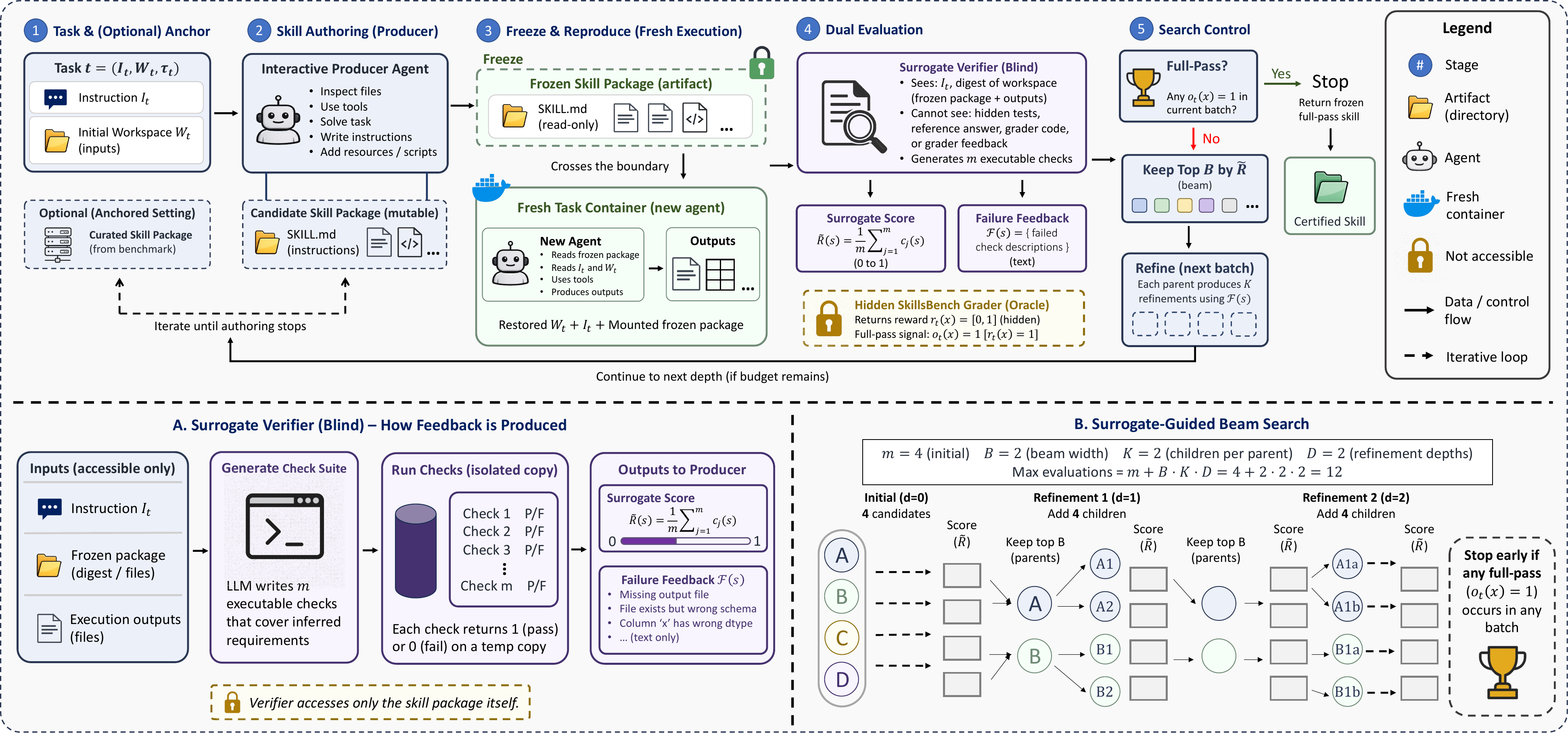}
  \caption{\textbf{reSolve evolves a skill package through a five-stage
  solve-and-reproduce loop.}  The top row follows one candidate through
  stages~1--5, which are task and optional curated anchor, interactive authoring, frozen
  re-execution by a new agent, dual evaluation (surrogate score and hidden
  grader), and search control.  A full pass returns the frozen skill.  Otherwise,
  the beam keeps the top candidates under $\widetilde R$ and refines them.
  Panel~A details how the blind verifier generates and runs executable checks,
  and Panel~B illustrates the search configuration ($m{=}4$, $B{=}2$, $K{=}2$,
  $D{=}2$, at most 12 evaluations).}
  \label{fig:arch}
\end{figure*}

\Cref{fig:arch} organizes this process into five stages.  Stage~1 combines the
task instruction and workspace with an optional curated anchor.  Stage~2 runs
an interactive producer that authors a mutable candidate skill package.  In
Stage~3, the package is frozen and crosses the reproduction boundary to a fresh
container, where a new agent uses it to execute the task.  Stage~4 applies dual
evaluation, in which a surrogate verifier produces a score and textual failure
feedback while the hidden grader independently makes the full-pass decision.  Stage~5
either returns a passing skill or retains the most promising candidates for
another round of refinement.  Panel~A details the surrogate verification
process and Panel~B illustrates the beam-search configuration.  We explain
these components in the same order below.

\subsection{Problem setup}\label{sec:setup}

A task is $t=(I_t,W_t,r_t)$.  Here, $I_t$ is the natural-language
instruction, $W_t$ is the initial workspace containing the task inputs, and
$r_t(x)\in[0,1]$ is the hidden grader's reward for the files $x$
produced by an agent.  Because some tasks award partial credit, reSolve distinguishes
between a nonzero score and a complete solution.  It treats
\begin{equation}
  o_t(x)=\mathbb{1}[r_t(x)=1]
  \label{eq:oracle-gate}
\end{equation}
as the full-pass signal, where $o_t(x)=1$ means that the candidate has satisfied
the grader completely.  The producer never sees hidden tests, reference answers,
grader code, or grader-side diagnostic text.

A candidate skill is a frozen directory containing a \texttt{SKILL.md} file and
any supporting instructions, resources, or scripts that the agent has chosen to
include.  To use the skill, a fresh agent reads this directory together with the
original task instruction and workspace, then produces the files that the grader
will inspect.  Freezing the package does not freeze this later trajectory.  Two
agents can interpret the same instructions differently, invoke different tools,
or recover from errors in different ways.

This is the reproduction gap in concrete terms.  A candidate may pass once
during evolution yet score lower when fresh agents use it again.  reSolve does not
eliminate stochasticity from skill use; it instead requires every candidate to
cross the same fresh-agent boundary before a pass is credited.

\subsection{Solve-and-Reproduce}\label{sec:snr}

Solve-and-reproduce separates the agent that writes a skill from the agent that
uses it.  During authoring, the producer receives the task instruction, the
initial workspace, and the available environment documentation.  In the
anchored setting, it may also read and execute the benchmark's curated skill
package.  The from-scratch ablation withholds that package.  In both settings,
the producer can inspect files, call tools, and solve the task interactively,
then package its findings into a candidate skill.

Once authoring ends, reSolve freezes the candidate directory.  Only this directory
crosses the reproduction boundary, while the producer's conversation, intermediate
reasoning, tool history, and modified authoring workspace are discarded.  reSolve
then restores the original task workspace in a fresh container, mounts the
frozen package, and asks a new agent to carry out the task.  The hidden grader is
called only after this independent execution has produced its output files.

Let $\pi_p$ denote the producer and $\pi_d$ the deployment agent.  Both
instantiate the same underlying model but operate in disjoint contexts.  The
producer interacts with the environment and distills its experience into a
candidate,
\begin{equation}
  s \sim \pi_p(\cdot \mid I_t, W_t, s^*),
  \label{eq:produce}
\end{equation}
where $s^*$ is the curated skill package, while a
fresh deployment agent produces output files from the frozen package alone,
\begin{equation}
  x \sim \pi_d(\cdot \mid I_t, W_t, s).
  \label{eq:deploy}
\end{equation}
$\pi_d$ conditions on $s$ but not on the producer's trajectory or
intermediate workspace, so the frozen package is the sole channel between the two
agents.

What is tested during evolution therefore matches what is deployed, namely the
frozen package rather than the authoring trace.  Standard skill evolution
certifies a joint trajectory in which one agent both writes $s$ and solves the
task.  reSolve instead certifies that $o_t(x)=1$ for an independently
sampled~$\pi_d$, which is the condition required at deployment.  Because $\pi_d$
is stochastic, a single evolution-time pass does not guarantee future success,
and the repeated deployment trials in \Cref{sec:exp-setup} quantify the
remaining variation.

\subsection{Blind surrogate verifier}\label{sec:surrogate}

The hidden grader can determine whether an execution is correct, but it gives
the producer little help in repairing a failed skill.  reSolve therefore creates a
surrogate check suite for each evaluated candidate.  The verifier receives the
task instruction and a digest of the accessible candidate workspace, including
the frozen package and the files produced during fresh execution.  It cannot
access the hidden tests, reference answer, grader implementation, or grader-side
feedback.

From this information, the verifier writes $m$ executable checks that cover the
requirements it can infer from the task description and workspace.  These
include, for instance, checks that an expected output file exists and conforms
to the prescribed format, that numerical outputs fall within physically
reasonable ranges, or that required columns appear in a result table.  Each check returns one when it passes and
zero when it fails.  reSolve summarizes their results as
\begin{equation}
  \widetilde R(s)=\frac{1}{m}\sum_{j=1}^{m}c_j(s),
  \qquad
  \mathcal F(s)=\{\text{descriptions of failed checks}\}.
  \label{eq:surrogate}
\end{equation}
$\widetilde R(s)$ scores candidate progress, and $\mathcal F(s)$ tells the
producer what to repair, for example a missing output file versus one with the
wrong schema or content.  Refinement receives only these surrogate descriptions,
never hidden-grader diagnostics.

The generated checks run on a temporary copy of the candidate workspace, so
their writes do not affect the files later sent to the grader.  reSolve generates
one fresh suite per candidate rather than fixing a single suite for the entire
task.  Different candidates therefore face slightly different checks, which
broadens coverage of inferred requirements at the cost of strict cross-candidate
comparability.  Accordingly, $\widetilde R$ is a ranking heuristic within
the beam, not a proxy for the grader reward.

\subsection{Surrogate-guided beam search}\label{sec:search}

A single refinement chain can commit early to an unpromising candidate.  reSolve
instead represents evolution as a solution-construction graph.  Each node is an
evaluated skill package, and each edge is a refinement that rewrites a parent
using its surrogate failure descriptions.  Beam search explores several paths
through this graph while keeping the evaluation budget explicit.

The search begins by sampling $m$ initial skill packages from the producer at
varying temperatures.  Each package is independently authored, so the producers
share the same task context but do not observe one another's outputs.  Every package is frozen and
fresh-executed using the procedure in \Cref{sec:snr}, then scored by both the
surrogate and the hidden grader.  If any candidate receives full reward, reSolve
retains a full-pass package and stops after the current concurrent batch has
completed.  Otherwise, it keeps the $B$ non-passing candidates with the highest
surrogate scores $\widetilde R$.

Each retained parent $s$ then produces $K$ independently sampled refinements.
The refiner receives the parent's frozen package, the failed surrogate checks
$\mathcal{F}(s)$ with their textual descriptions, and the original task context
$(I_t, W_t)$.  It revises the skill in response to the diagnosed failures, for
example by restructuring the package layout or rewriting a script that produced
malformed output.  Parents remain eligible for the next beam, so a child whose
surrogate score falls below its parent is pruned rather than replacing it.

reSolve repeats this refinement step for $D$ depths.  The total evaluation budget
is at most $m + B \cdot K \cdot D$ candidates, each requiring one fresh
execution and one surrogate assessment, and early stopping on a full pass may
use fewer.  With the reported configuration ($m{=}4$, $B{=}2$, $K{=}2$,
$D{=}2$), this gives at most twelve evaluations per task.  Search breadth and
depth thus expose a test-time-compute knob without changing the underlying
model.  If the budget is exhausted without a full pass, no package is
certified, and the run retains the highest-scoring candidate as a
fallback. Curated package (if any) is always included in the initial beam, so the fallback is at least as good as the curated baseline. This is noted as the keep-better rule.

The full-pass signal tells the search whether to continue but not why a
candidate failed.  Repair guidance comes entirely from $\mathcal{F}$, the
surrogate's independently generated feedback.

\begin{algorithm}[t]
\caption{reSolve self-evolution for one task}
\label{alg:resolve}
\begin{algorithmic}[1]
\REQUIRE task $t$; initial width $m$; beam width $B$; children $K$; depth $D$
\STATE $C\leftarrow\{\textsc{ProposeSkill}(t;\tau_i)\}_{i=1}^{m}$
\STATE fresh-execute and evaluate every skill $s\in C$
\IF{a skill in $C$ receives full reward}
  \STATE \textbf{return} a frozen full-pass skill
\ENDIF
\STATE $\mathcal B\leftarrow$ the top $B$ skills in $C$ by surrogate score
\FOR{$d=1$ \TO $D$}
  \STATE $C'\leftarrow\bigcup_{s\in\mathcal B}
    \{\textsc{RefineSkill}(s,\mathcal F(s);\tau_{d,j})\}_{j=1}^{K}$
  \STATE fresh-execute and evaluate every skill $s'\in C'$
  \IF{a skill in $C'$ receives full reward}
    \STATE \textbf{return} a frozen full-pass skill
  \ENDIF
  \STATE $\mathcal B\leftarrow$ the top $B$ skills in $\mathcal B\cup C'$ by surrogate score
\ENDFOR
\STATE \textbf{return} the recorded fallback, or no package if none was produced
\end{algorithmic}
\end{algorithm}

\section{Experiments}\label{sec:exp}

To evaluate reSolve, we run the full framework on 86
SkillsBench tasks with DeepSeek-V4-Pro and measure main results, ablation effects, and
execution stability.  We then repeat the pipeline with Gemma~4 31B to test whether the
gains transfer to a substantially weaker, non-reasoning model.

\subsection{Experimental setup}\label{sec:exp-setup}

\paragraph{Benchmark.}
We evaluate on SkillsBench \citep{skillsbench}, a benchmark of 87 coding tasks
spanning eight official categories, including 14 Natural Science tasks that cover
astronomy, seismology, hydrology, materials science, physics, and biomedical analysis.
Each task ships with a curated skill package and an executable grader.  We use the 86
tasks whose resources are network-public and exclude from every condition the one task
that requires withheld network access.

\paragraph{Model.}
Our primary model is DeepSeek-V4-Pro, a Mixture-of-Experts (MoE) model with 1.6T total parameters and 49B activated parameters, accessed through DeepSeek's official API with high
reasoning effort.  All main results in this paper are reported on this model.  As a generalization test, we additionally run the full pipeline with Gemma~4 31B,
a 31-billion-parameter non-reasoning model, to examine whether reSolve's gains transfer to
a substantially weaker base model.

\paragraph{Evaluation conditions.}
\emph{No skill} runs the agent without any task skill.  \emph{reSolve Curated} runs the
model on our harness with the human-written SkillsBench skill package mounted.
\emph{reSolve Self-Evolved} is the output of the full framework.  For each task, reSolve
evolves a candidate skill package and keeps whichever of the evolved and curated
conditions performs better (the keep-better rule, \Cref{sec:search}).  The no-skill and reSolve
Curated conditions are paired baselines, and reSolve Self-Evolved is our primary
result.

\paragraph{Deployment and missing trials.}
Each deployed condition runs three independent trials per task, each with a budget of
1{,}800\,s, 80 agent steps, and a 120\,s per-command timeout.
During the evolution stage, each candidate skill is evaluated with a fixed budget of
1{,}200\,s.  The per-task metrics are
\begin{equation}
  \bar r_t = \tfrac{1}{3}\sum_{j=1}^{3} r_{t,j},
  \qquad
  \operatorname{pass@3}_t=\mathbb{1}\!\bigl[\max_{j} r_{t,j}=1\bigr].
  \label{eq:metric}
\end{equation}
We refer to $\bar r_t$ as the \emph{mean-of-3} score.  The macro metrics average these
values over 86 tasks.

\paragraph{Uncertainty.}
All confidence intervals are computed with a paired nonparametric bootstrap
with 100{,}000 resamples.  In each resample, we draw 86 tasks with replacement
and recompute the aggregate metric. When comparing two conditions, we draw the same
task indices for both, preserving the paired structure so that per-task difficulty
differences cancel out.  These
intervals reflect sampling variation over the task set, i.e., how results might shift if
a different set of tasks were evaluated. 

\subsection{Main results}\label{sec:exp-main}

\paragraph{Leaderboard.}
\Cref{tab:main} places the reSolve results alongside all 24 model--agent configurations in
the SkillsBench v1.1 official leaderboard \citep{skillsbench}.  Frontier models occupy the top of the
leaderboard.  GPT-5.5/OpenHands leads with $67.3\%$
mean-of-3 with curated skills, followed by GPT-5.5/Codex ($66.5\%$) and Opus~4.7/Claude
Code ($61.2\%$).  Both of our models sit in the lower tier.  DeepSeek-V4-Pro ranks 14th
at $50.1\%$, and Gemma~4 31B's no-skill score of $14.7\%$ would place it second to last
among all entries.  Running the two models on reSolve with no skill yields $30.6\%$ and
$14.7\%$, and mounting the curated package (reSolve Curated) raises them to $60.1\%$ and
$36.2\%$, comparable to the 5th- and 21st-ranked official entries.  Self-evolution (reSolve
Self-Evolved) then lifts DeepSeek-V4-Pro to $74.9\%$, $7.6$ points above the strongest official
result, and Gemma~4 31B to $41.4\%$, comparable to GPT-5.4 Mini.

\begin{table*}[t]
\centering
\small
\caption{\textbf{Main result in complete SkillsBench v1.1 context.}\label{sec:exp-leaderboard}
Scores and $\Delta$ are percentages. Background colors mark the
\colorbox{rankfirst}{\strut first}, \colorbox{ranksecond}{\strut second}, and
\colorbox{rankthird}{\strut third} best value per official column, and ties share a color.}
\label{tab:main}
\providecolor{rankfirst}{HTML}{FFF2CC}
\providecolor{ranksecond}{HTML}{E9EBEE}
\providecolor{rankthird}{HTML}{FBE2CB}
\providecolor{seedtint}{HTML}{EDF6F4}
\begin{tabular}{@{}rll rrr@{}}
\toprule
Rank & Model & Agent & No skills & Curated & $\Delta$ \\
\midrule
\multicolumn{6}{@{}l}{\emph{SkillsBench Leaderboard}} \\
\midrule
1 & GPT-5.5 & OpenHands & \cellcolor{rankfirst}51.5 & \cellcolor{rankfirst}67.3 & +15.8 \\
2 & GPT-5.5 & Codex & \cellcolor{ranksecond}46.8 & \cellcolor{ranksecond}66.5 & +19.7 \\
3 & Opus 4.7 & Claude Code & 43.0 & \cellcolor{rankthird}61.2 & +18.2 \\
4 & Gemini 3.1 Pro & Gemini CLI & 36.0 & 60.8 & \cellcolor{rankthird}+24.8 \\
5 & GLM 5.1 & OpenHands & 32.7 & 58.4 & \cellcolor{rankfirst}+25.7 \\
6 & Gemini 3 Flash & Gemini CLI & 34.2 & 54.6 & +20.4 \\
7 & Opus 4.8 & OpenHands & \cellcolor{rankthird}45.7 & 54.1 & +8.4 \\
8 & Kimi K2.6 & OpenHands & 33.4 & 54.0 & +20.6 \\
9 & Opus 4.7 & OpenHands & 42.1 & 53.1 & +11.1 \\
10 & MiniMax M3 & OpenHands & 29.7 & 53.0 & +23.3 \\
11 & Gemini 3.1 Pro & OpenHands & 33.8 & 52.8 & +19.0 \\
12 & GPT-5.2 & Codex & 29.7 & 51.7 & +22.0 \\
13 & Opus 4.6 & Claude Code & 33.7 & 50.2 & +16.5 \\
14 & DeepSeek V4 Pro & OpenHands & 26.9 & 50.1 & +23.2 \\
15 & Opus 4.5 & Claude Code & 23.8 & 49.0 & \cellcolor{ranksecond}+25.2 \\
16 & Gemini 3.5 Flash & OpenHands & 41.1 & 48.2 & +7.1 \\
17 & Sonnet 4.6 & OpenHands & 33.5 & 47.2 & +13.6 \\
18 & DeepSeek V4 Flash & OpenHands & 27.5 & 44.7 & +17.2 \\
19 & Grok 4.3 & OpenHands & 22.8 & 41.7 & +18.8 \\
20 & GPT-5.4 Mini & OpenHands & 29.9 & 41.4 & +11.5 \\
21 & Sonnet 4.5 & Claude Code & 16.7 & 36.2 & +19.5 \\
22 & MiniMax M2.7 & OpenHands & 18.1 & 34.9 & +16.8 \\
23 & Haiku 4.5 & Claude Code & 8.8 & 30.1 & +21.3 \\
24 & Gemini 3.1 Flash Lite & OpenHands & 16.0 & 20.1 & +4.1 \\
\midrule
\multicolumn{6}{@{}l}{\emph{reSolve Results}} \\
\midrule
\rowcolor{seedtint} -- & DeepSeek V4 Pro & reSolve Curated & 30.6$^{\,14th}$ & 60.1$^{\,5th}$ & +29.6 \\
\rowcolor{seedtint} -- & DeepSeek V4 Pro & \textbf{reSolve Self-Evolved} & 30.6$^{\,14th}$ & \textbf{74.9}$^{\,1st}$ & +44.3 \\
\rowcolor{seedtint} -- & Gemma 4 31B & reSolve Curated & 14.7$^{\,24th}$ & 36.2$^{\,21st}$ & +21.5 \\
\rowcolor{seedtint} -- & Gemma 4 31B & \textbf{reSolve Self-Evolved} & 14.7$^{\,24th}$ & \textbf{41.4}$^{\,20th}$ & +26.7 \\
\bottomrule
\end{tabular}

\end{table*}

\begin{table}[H]
\centering
\small
\caption{\textbf{reSolve aggregate.} Entries are percentages.  Confidence
intervals are task-bootstrap intervals for the mean, and differences are paired against
curated.}
\label{tab:headline}
\begin{tabular}{lrrrr}
\toprule
Condition & Mean-of-3 & 95\% CI & $\Delta$ curated & Pass@3 \\
\midrule
No skill & 30.6 & [23.1, 38.4] & -29.6 & 45.3 \\
Curated skill & 60.1 & [51.2, 68.8] & +0.0 & 68.6 \\
\textbf{reSolve Self-Evolved} & \textbf{74.9} & [66.3, 83.0] & +14.8 & 79.1 \\
\bottomrule
\end{tabular}

\end{table}

\begin{figure}[t]
\centering
\includegraphics[width=0.60\linewidth]{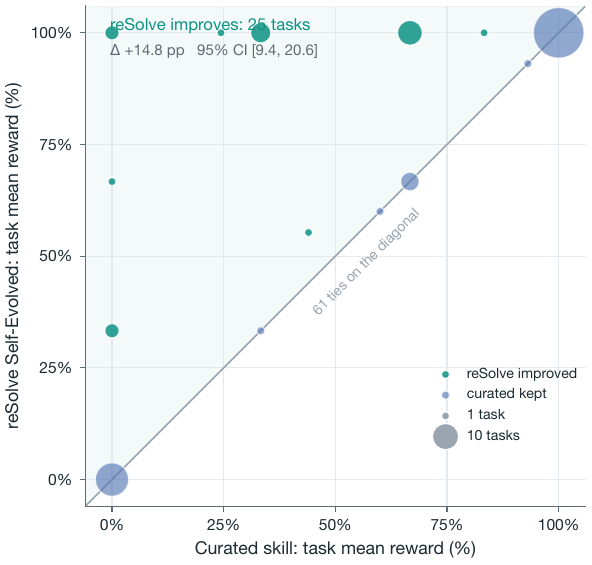}
\caption{\textbf{25 tasks yield improvements on reSolve.} Each marker compares
the reSolve Curated and reSolve Self-Evolved conditions.  Tasks with identical score pairs are
aggregated, with marker area proportional to the task count.  Points above the diagonal
represent tasks where reSolve found a better skill.}
\label{fig:main}
\end{figure}

\paragraph{Effect of self-evolution.}
Self-evolution delivers a clear improvement on DeepSeek-V4-Pro (\Cref{tab:headline}).
The mean-of-3 score rises from $60.1\%$ (reSolve Curated) to $74.9\%$, a paired difference
of $+14.8$ points with a $95\%$ bootstrap interval of $[9.4, 20.6]$, and pass@3 rises
from $68.6\%$ to $79.1\%$.  At the task level (\Cref{fig:main}), self-evolution improves
25 of 86 tasks.  The gains are also heavy-tailed.  Rather than small uniform shifts
across the benchmark, most of the improvement comes from a subset of tasks that improve
substantially, which appear in \Cref{fig:main} as points far above the diagonal.
The composition of these gains shows where the search helps.  Of the 25 improved tasks,
15 start from a curated score below $50\%$, and 20 reach a perfect score across all
three deployment trials after evolution.  The 11 tasks that gain at least $50$ points
account for roughly two thirds of the total improvement.  Self-evolution therefore acts
less as uniform polish and more as a rescue mechanism.  It converts tasks where the
curated package fails or passes only intermittently into reliably passing ones.  This
pattern is consistent with the search objective, which terminates only on a full pass
and thus steers candidates toward complete solutions rather than incremental partial
credit.

\paragraph{Cross-model generalization.}
The Gemma~4 31B rows in \Cref{tab:main} test whether these gains depend on base-model
capability.  Without skills Gemma reaches $14.7\%$, roughly half the DeepSeek baseline.
Curated skills lift it to $36.2\%$, and self-evolution adds $+5.2$ points for $41.4\%$.
The gains are markedly more modest than for DeepSeek.  Evolution improves only 10 of 86
tasks compared with 25 for DeepSeek, and in leaderboard terms DeepSeek climbs from 14th
to 1st while Gemma moves from 24th to 20th.  The gap holds in relative terms as
well, since $+5.2$ points on a $36.2\%$ baseline is a smaller proportional gain than
$+14.8$ on $60.1\%$.  This asymmetry suggests that the benefit of self-evolution is
bounded by the capability of the generator itself.  Evolving a skill requires the
producer to interpret execution feedback, diagnose why a candidate failed, and
synthesize a revised package, and a weaker model succeeds at these steps less often.
Self-evolution therefore amplifies existing capability rather than compensating for its
absence, and we expect the benefit to shrink further as the base model weakens.

\FloatBarrier

\subsection{Ablation studies}\label{sec:exp-abl}

We then compare the full reSolve framework with three ablation settings on the same 86 tasks
(\Cref{tab:ablation}).  Each ablation removes one component and follows the same
evaluation protocol where applicable.  The \emph{w/o
surrogate ranking} setting removes the surrogate score $\widetilde R$ from beam
selection, so the search can no longer prioritize candidates by inferred progress when
choosing which ones to retain and refine.  The \emph{w/o beam search} setting replaces
the beam with greedy search, which follows a single refinement chain and keeps only the
current best candidate instead of maintaining multiple parents. The \emph{w/o curated skill package} setting
evolves from scratch, without access to the curated skill package.  Each setting contains
one stochastic evolution run followed by three attempted deployment trials.  The
intervals resample tasks within those realized runs and do not capture run-to-run search
variance.

\begin{table}[H]
\centering
\small
\caption{\textbf{Ablation effects relative to the full reSolve framework.} Entries are mean
reward percentages and paired task-bootstrap intervals.}
\label{tab:ablation}
\begin{tabular}{lrrr}
\toprule
Condition & Mean-of-3 & Effect vs. full & 95\% CI \\
\midrule
Full reSolve & 74.9 & +0.0 & [0.0, 0.0] \\
w/o surrogate ranking & 66.0 & -8.9 & [-14.3, -4.5] \\
w/o beam search (greedy only) & 59.4 & -15.5 & [-20.3, -11.3] \\
w/o curated skill package & 44.0 & -30.9 & [-40.7, -21.6] \\
\bottomrule
\end{tabular}

\end{table}

\begin{figure}[H]
\centering
\includegraphics[width=0.90\linewidth]{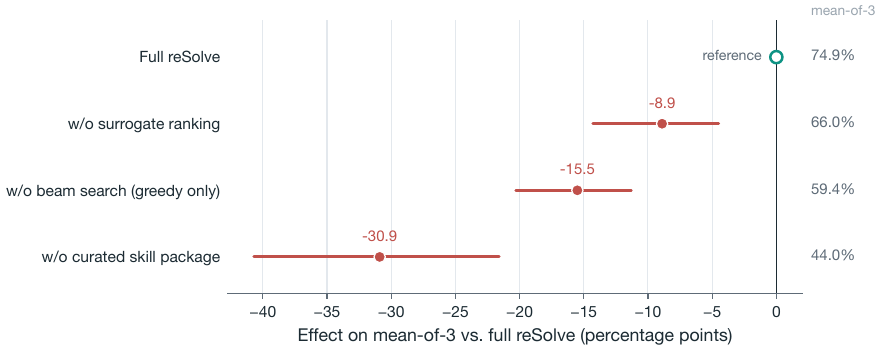}
\caption{\textbf{Paired ablation effects with 95\% task-bootstrap intervals.} Zero is the
full reSolve framework.  These intervals describe heterogeneity across tasks for one
evolution realization per arm.}
\label{fig:ablation}
\end{figure}

\paragraph{Refinement is more effective than generation from scratch.}
Without the curated package, reSolve reaches $44.0\%$ mean reward,
$30.9$ points below the full framework ($95\%$ CI $[-40.7,-21.6]$).  This corroborates
our position in \Cref{sec:intro} that self-evolution is most effective as refinement
rather than generation from scratch.  Evolving without an expert anchor places the
entire burden of exploration on the generator, which must discover the task structure,
search the solution space, and distill reliable procedures on its own.  Doing this well
demands a substantially more capable, and therefore more expensive, generator.  Anchored
on the curated package, the producer instead starts from working expert knowledge and
only needs to find a task-specific improvement.  That said, from-scratch evolution is
far from ineffective.  It lifts the model from the $30.6\%$ no-skill baseline to
$44.0\%$, a $+13.4$-point gain achieved without any human-written skill.  Placed
against the official curated-skill column in \Cref{tab:main}, this score reaches the
lower middle of the leaderboard, slightly above Grok~4.3 ($41.7\%$) and GPT-5.4~Mini
($41.4\%$) running with human-curated packages.  Fully autonomous evolution can
therefore already match what human curation delivers to mid-tier models, while
refinement remains the far more economical path in this configuration.

\paragraph{Beam search converts additional compute into coverage.}
Greedy search reaches $59.4\%$ mean reward, a paired effect of $-15.5$ points
($95\%$ CI $[-20.3,-11.3]$).  This is consistent with the proposed role of beam search,
in which independent initial attempts and multiple retained parents cover solutions
that a single refinement chain misses.  Because breadth and depth are explicit
parameters, a larger evaluation budget directly widens the explored region of the
solution-construction graph, so the search trades test-time compute for a better chance
of containing a reproducible solution.  A greedy chain offers no such trade.  It
commits to its first candidate, has no remaining diversity to recover from a flawed
start, and propagates the cost of that start through every subsequent refinement.

\paragraph{Surrogate ranking steers the refinement budget.}
Disabling surrogate ranking yields $66.0\%$ mean reward, a paired effect of $-8.9$
points ($95\%$ CI $[-14.3,-4.5]$).  The interval excludes zero, so the ranking
contributes signal beyond sampling breadth alone.  The retained parents determine where
the entire refinement budget flows, and without $\widetilde R$ the beam cannot
distinguish promising from unpromising candidates, so part of that budget is spent
refining the wrong parents.  The effect remains the smallest of the three though, which is
consistent with the surrogate acting as a selection heuristic inside a search whose
breadth already covers much of the candidate space.  We expect its role to grow with
wider beams, where selecting the right parents rather than raw coverage becomes the
binding constraint.  That growth calls for care.  One fresh LLM-written suite per
candidate remains a noisy value function, and stronger optimization against an
imperfect proxy can eventually reduce the target score rather than improve it
\citep{rewardoveroptimization}.

\FloatBarrier

\subsection{Further analyses}\label{sec:exp-analysis}

\subsubsection{Execution stability}\label{sec:exp-det}

\begin{figure}[H]
\centering
\includegraphics[width=0.82\linewidth]{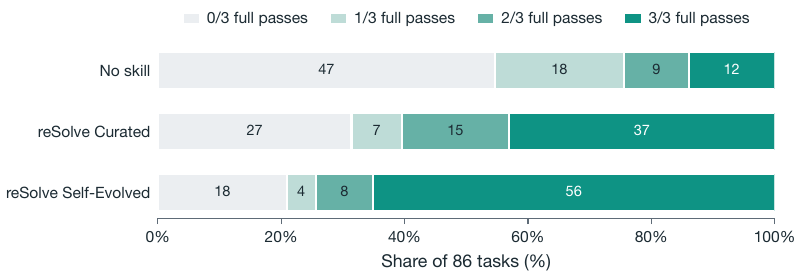}
\caption{\textbf{Distribution of full passes across three deployment trials.}
Each bar partitions the 86 tasks by how many of the three trials end in a full pass.
The reSolve Self-Evolved row applies the
keep-better rule of \Cref{sec:search}; the evolved-package and direct-artifact rows
deploy each packaging variant unconditionally.}
\label{fig:stability}
\end{figure}

\begin{figure}[H]
\centering
\includegraphics[width=0.82\linewidth]{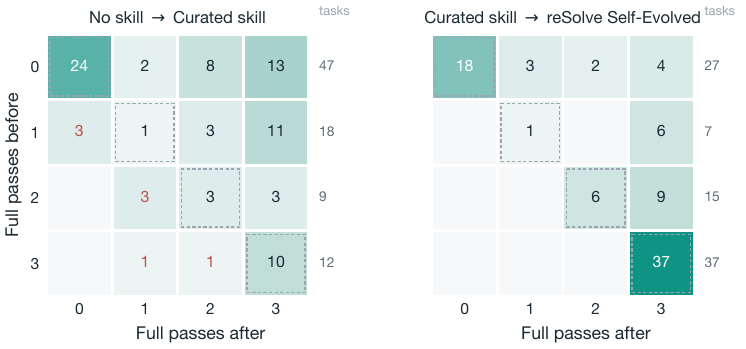}
\caption{\textbf{Task-level movement of full-pass counts along the condition ladder.}
Cells count tasks by full passes before (rows) and after (columns) each step; dashed
diagonal cells are unchanged, and red counts below the diagonal mark tasks that lose
passing trials.  Curated rescue concentrates above the diagonal of the left panel,
and the keep-better rule leaves no cell below the diagonal in the right panel.
Row sums appear in the right margin of each panel.}
\label{fig:stability-flow}
\end{figure}

A frozen skill package does not freeze its deployment trajectory.  The deployment agent
re-derives the solution on every trial, choosing its own commands, file edits, and
recovery steps, so the same package can yield different rewards across the three
trials.  We therefore treat stability as a measured property of each condition and
analyze it in three steps.  We report the per-task distribution of full passes
(\Cref{fig:stability}), a slot-level decomposition of every non-passing trial
(\Cref{tab:stability}), and the task-level movement between conditions
(\Cref{fig:stability-flow}).

\Cref{fig:stability} counts, for each condition, how many of a task's three deployment
trials end in a full pass.  Reliability rises along the condition ladder.  The number
of tasks that pass all three trials grows from 12 of 86 with no skill to 37 with reSolve
Curated and 56 with reSolve Self-Evolved, and full-pass trial slots grow from 72 to 148 to
188 of 258.

\paragraph{Instability is dominated by silent semantic divergence.}
\Cref{tab:stability} assigns each of the 258 intended trial slots to one of three
disjoint outcomes, namely a full pass, a graded failure in which the trial completes without
any flagged error and the grader scores it below 1, and a flagged execution error such
as an agent timeout or a verifier crash.  Graded failures dominate every condition.
Under reSolve Curated, 93 of the 110 non-passing slots are graded failures against 17
flagged errors, and under reSolve Self-Evolved the split is 57 against
13.  The proportions are similar without any skill (160 of 186), so the pattern
is a property of the executor rather than of the packages.  The agent rarely crashes;
it completes confidently and produces files the grader rejects.  The traces of
\texttt{gravitational-wave-detection} illustrate the mechanism.  Its two passing trials
locate the mounted package, read the skill files, and implement the packaged recipe as
one consolidated script.  The failing trial finds the same package and the same
library, but then explores several waveform approximants, assembles a results file by
hand from separate runs, and lands on values the grader rejects.  The package told the
agent what to do, yet it left the analysis decisions to the executor, and those
decisions vary across samples.  A second mechanism is budget truncation.  On
\texttt{paratransit-routing} and \texttt{seismic-phase-picking} every trial ends with a
flagged timeout, yet both tasks pass in two of three slots because the grade depends on
whatever output the agent had already written when the budget expired; this is also why
a flagged trial can still count as a full pass in \Cref{tab:stability}.  Failing trials
do not reflect less effort either, as the trace audit in \Cref{sec:exp-fail} shows that
they tend to use \emph{more} tool steps than passing trials of the same task.

\begin{table}[H]
\centering
\small
\setlength{\tabcolsep}{4.5pt}
\caption{\textbf{Slot-level decomposition of the $86\times3=258$ deployment trials.}
A slot is a full pass when its reward reaches 1, even if the runner also flagged an
error.  A graded fail is a trial that completes with no flagged error and scores below 1.
An exec.\ error is a flagged agent, environment, or verifier error.
Mixed tasks pass one or two of their three trials.}
\label{tab:stability}
\begin{tabular}{lrrrr}
\toprule
 &  & \multicolumn{2}{c}{Non-passing slots} &  \\
\cmidrule(lr){3-4}
Condition & Full pass & Graded fail & Exec.\ error & Mixed tasks \\
\midrule
No skill & 72 & 160 & 26 & 27 \\
Curated skill & 148 & 93 & 17 & 22 \\
Evolved package only & 169 & 76 & 13 & 21 \\
Direct artifact (no agent) & 177 & 76 & 5 & 3 \\
\textbf{reSolve Self-Evolved} & 188 & 57 & 13 & 12 \\
\bottomrule
\end{tabular}

\end{table}

\paragraph{Self evolution consolidates intermittent tasks.}
\Cref{fig:stability-flow} tracks each task's full-pass count across the condition
ladder and separates the two contributions.  Moving from no skill to reSolve Curated
mainly rescues unsolvable tasks.  Of the 47 tasks that never pass without a skill, 13
become reliable three-of-three passes and 10 more become intermittent, and 11 of the 18
tasks with a single lucky pass consolidate to three.  The curated package is not
uniformly safe, though, as 8 tasks lose at least one passing slot after mounting it
(the red counts below the diagonal).  Moving from reSolve Curated to reSolve Self-Evolved
instead mainly consolidates the intermittent band.  Of the 19 tasks that newly reach
three passes, 15 already passed once or twice under the curated package and only 4 come
from the never-passing band, and no task loses a passing slot, since the keep-better
rule retains the curated trials whenever evolution fails to improve on them.
Self-evolution is therefore better at making partially working procedures reliable than
at cracking tasks the curated package cannot touch at all.  This matches the
solve-and-reproduce objective.  A candidate is certified only when a fresh agent
reproduces the solution from the frozen package, so the search favors packages that
leave the executor little room for interpretation, typically by shipping an executable
entry point instead of prose the agent must re-derive (\Cref{sec:exp-fail}).

\FloatBarrier

\subsubsection{Domain-level success rates}\label{sec:exp-cat}

\begin{figure}[H]
\centering
\includegraphics[width=\linewidth]{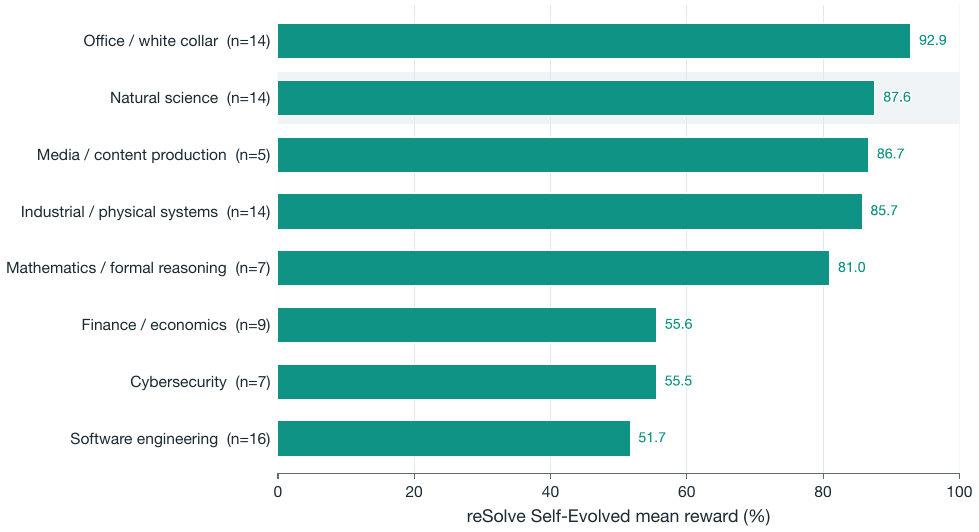}
\caption{\textbf{reSolve Self-Evolved mean reward by official SkillsBench category.}
Values are percentages under the primary non-null convention.  Natural Science tasks
are highlighted, and category samples are small ($n=5$--$16$).}
\label{fig:categories}
\end{figure}

\Cref{fig:categories} reports the reSolve Self-Evolved mean reward on each of SkillsBench's
eight official categories.  The agent is strongest on office and white-collar work
($92.9\%$), natural science ($87.6\%$), media and content production ($86.7\%$),
industrial and physical systems ($85.7\%$), and mathematics and formal reasoning
($81.0\%$).  These categories share a common task shape, a well-specified deliverable,
such as a formatted document, a data product, or a simulation output, that a scripted
pipeline can produce once the correct procedure is known.  This is the regime skill
packages are designed for, and where shipping a verified, re-executable procedure pays
off most.  Mean reward is lower on finance and economics ($55.6\%$), cybersecurity
($55.5\%$), and software engineering ($51.7\%$), where each task demands more
open-ended exploration, such as navigating an unfamiliar codebase or probing an
adversarial environment, so less of the work can be captured ahead of time as a
reusable procedure.

\paragraph{Natural Science.}
Scientific work is an intended application of this research, so we note the 14 official
Natural Science tasks separately.  The subset covers astronomy, seismology, hydrology,
materials, physical simulation, and biomedical analysis.  reSolve Self-Evolved reaches
$87.6\%$ mean reward here, the second-highest category and well above the $74.9\%$
overall average.  Scripted analysis pipelines over scientific data, with a concrete
checkable output, are exactly where the deployed system is most reliable.

\FloatBarrier

\subsubsection{How agents consume skills, and what that implies for skill creation}\label{sec:exp-fail}

The preceding sections evaluate skills by the reward they produce.  This section asks a
different question.  What does the deployed agent actually do with a skill, and
what does the answer imply for how skills should be created?  Our starting observation
is that a frozen package is not executed by a fixed interpreter.  It conditions a
stochastic tool-using policy that must discover the package, ground its instructions in
the current filesystem, choose between invoking and rewriting the bundled code,
validate intermediate results, and decide when to stop.  Each stage fails for a
different reason and calls for a different remedy, a decomposition consistent with
agent evaluations that separate planning, tool grounding, state tracking, and
completion rather than folding success into a single reasoning variable
\citep{taubench,travelplanner,toolsandbox,agentboard}.  We audit the archived
deployment traces stage by stage and read them as evidence about where a skill's value
is created and where it is destroyed.

\paragraph{A frozen package does not fix the behavior it induces.}
Among the 85 tasks with at least two observed rewards, 20 ($23.5\%$) change reward
across trials of the same package in the same environment, which echoes the
repeated-rollout inconsistency reported for interactive tool agents \citep{taubench}.
The unit that carries reliability is therefore not the package but the
package-executor pair, and pass@3 measures whether a solution is recoverable somewhere in
the trial budget, not the probability that the next execution succeeds.  For skill
creation this reframes the objective.  A skill is not finished when one trajectory
passes; it is finished when it removes enough decisions from the executor that the
induced behavior distribution concentrates on the solution.

\paragraph{Failed trials are not lazy trials.}
Fourteen tasks contain both a full-pass and a zero-reward trace with no recorded runner
error.  On 11 of the 14, the failed trace uses more tool steps, with a mean
within-task difference of $+7.1$ steps.  Long trajectories do not cause failure, since
difficult states elicit more recovery attempts, but the pattern rejects the simplest
explanation, insufficient effort, and replaces it with a more consequential one, that extra
actions are not free.  As the execution-policy row of \Cref{tab:failure-taxonomy}
shows, an agent that keeps acting after the bundled solver has already produced a
correct artifact can overwrite that artifact.  Progress under an interpreting executor
is not monotone unless the harness makes it so.

\begin{table}[H]
\centering
\small
\setlength{\tabcolsep}{4pt}
\caption{\textbf{Failure modes.} ``Interpretation'' is derived from saved commands and outputs. }
\label{tab:failure-taxonomy}
\begin{tabular}{>{\raggedright\arraybackslash}p{0.17\textwidth}
                >{\raggedright\arraybackslash}p{0.24\textwidth}
                >{\raggedright\arraybackslash}p{0.27\textwidth}
                >{\raggedright\arraybackslash}p{0.24\textwidth}}
\toprule
Layer & Observable evidence & Representative case & Agent-system implication \\
\midrule
Search output
  & No retained package for 9/86 tasks
  & ADA bathroom repair has rewards $[0,1,0]$, but all three are bare-agent runs
  & Distinguish ``no package produced'' from ``package failed to reproduce'' \\
Package discovery
  & Semantic path attempted but physical path not recovered
  & Earthquake plate calculation, where \texttt{find A \&\& find B} short-circuits before \texttt{/skills}; the two trials that find the solver pass
  & Expose a canonical path and machine-readable entrypoint \\
Execution policy
  & Bundled solver runs, then later commands replace its behavior
  & FJSP optimization, where direct execution passes; post-execution repair fails
  & Run the frozen entrypoint first and preserve its output as a rollback checkpoint \\
Execution noise
  & Same packaged solver is invoked but numerical outputs differ
  & HVAC control, where settling times $50.0$, $58.5$, and $77.5$ under sensor noise yield rewards $[1,1,0]$
  & Seed stochastic components and test a distribution, not one replay \\
Termination
  & Runner error and grader reward disagree
  & 5 of 13 error-flagged trials still receive full reward; some timed-out agents had already written valid outputs
  & Grade the latest checkpoint separately from executor termination status \\
\bottomrule
\end{tabular}
\end{table}

\paragraph{Reproduction forces the agent to save executable scripts.}
All 77 retained evolved packages contain at least one executable file (\texttt{.sh},
\texttt{.py}, \texttt{.js}, or \texttt{.mjs}); the archive contains no prose-only
package.  We interpret this as selection rather than accident.  A candidate is retained
only if a fresh agent reproduces the solution from the frozen package, and a procedure
externalized as code survives re-execution with far less interpretation than guidance a
new agent must re-derive.  Consistent with this reading, a separately produced
agent-free arm that invokes the saved entrypoint directly reaches $71.4\%$, within
$3.5$ points of the full $74.9\%$ system with no interpreting agent at all, and with
near-perfect trial agreement.

\paragraph{What makes a good skill package.}
These observations converge on a single design target: a good skill package minimizes
the interpretation left to its executor.  Concretely, it should expose a entry point that carries out the procedure, rather than merely describe the
decisions an agent must make.  The package should make its dependencies, paths, inputs, outputs, and completion criteria explicit in order to control stochastic components where
possible. Execution should also be
idempotent and checkpointed, so that a valid intermediate artifact survives later
recovery attempts and can be graded independently of how the executor terminates.

Quality must therefore be assessed at the package boundary rather than inferred from
the trajectory that produced it. Under this criterion, a good skill is the smallest self-contained
interface that reliably reconstructs the successful outcome while leaving as few
consequential choices as possible to the executor. 

\section{Conclusion}\label{sec:conc}

In this work, we presented reSolve, a per-task, oracle-in-the-loop framework that lets an
LLM agent self-evolve its skills.  reSolve separates authoring from use.  Solve-and-reproduce
freezes each candidate package and certifies it only when a fresh agent reproduces the
solution in a new container.  A blind surrogate verifier adds the dense score and
failure descriptions that the hidden grader withholds, and verifier-guided beam search
exposes breadth and depth as an explicit test-time-compute parameter.  On 86 SkillsBench
tasks with DeepSeek-V4-Pro, reSolve Self-Evolved reached $74.9\%$ mean-of-3, $+14.8$ points
above the $60.1\%$ curated baseline on the same harness and $7.6$ points above the
strongest official curated-skill entry, though the official rows come from different
harnesses.  In a supplementary run, the recipe transferred to Gemma~4 31B with smaller
gains ($+5.2$ versus $+14.8$ points), so the benefit scales with the capability of the
model doing the evolving.  Ablations ranked the components.  The curated anchor
contributed most, then beam search, then surrogate ranking, and all three intervals
exclude zero.  From-scratch evolution still lifted the no-skill baseline from $30.6\%$
to $44.0\%$, within the range frontier models reach without skills, but it stayed well
below the anchored result.  Our trace audit located most remaining failures in the
interpreting executor rather than in the skill content, and the packages that survived
reproduction pressure converged on executable entry points instead of prose a new agent
must re-derive.  A pass certified by reSolve is one that search found and a fresh agent
then reproduced; that is the difference between a skill and a lucky trajectory.  Making
the executor as dependable as the packages it runs is the natural next step.

\bibliographystyle{bibstyle}
\bibliography{custom}

\end{document}